\documentclass[conference]{IEEEtran}
\IEEEoverridecommandlockouts
\usepackage{cite}
\usepackage{amsmath,amssymb,amsfonts}
\usepackage{algorithmic}
\usepackage{graphicx}
\usepackage{textcomp}
\usepackage{xcolor}
\usepackage{subfigure}
\graphicspath{{figures/}}

\def\BibTeX{{\rm B\kern-.05em{\sc i\kern-.025em b}\kern-.08em
		T\kern-.1667em\lower.7ex\hbox{E}\kern-.125emX}}
\begin{document}
\title{FasterRCNN Monitoring of Road Damages: Competition and Deployment\\}
	\author{\IEEEauthorblockN{Hascoet Tristan}
		\IEEEauthorblockA{\textit{Graduate School of System Informatics} \\
			\textit{Kobe University}\\
			Kobe, Japan \\
			tristan@people.kobe-u.ac.jp}
		\and
		\IEEEauthorblockN{Yihao Zhang}
		\IEEEauthorblockA{\textit{Graduate School of System Informatics} \\
			\textit{Kobe University}\\
			Kobe, Japan \\
			ryanzhang0901@gmail.com}
		\and
		\IEEEauthorblockN{Persch Andreas}
		\IEEEauthorblockA{\textit{Graduate School of System Informatics} \\
			\textit{Kobe University}\\
			Kobe, Japan \\
			a\_persch@web.de}
		\and
		\IEEEauthorblockN{Ryoichi Takashima}
		\IEEEauthorblockA{\textit{Graduate School of System Informatics} \\
			\textit{Kobe University}\\
			Kobe, Japan \\
			rtakashima@port.kobe-u.ac.jp}
		\and
		\IEEEauthorblockN{Tetsuya Takiguchi}
		\IEEEauthorblockA{\textit{Graduate School of System Informatics} \\
			\textit{Kobe University}\\
			Kobe, Japan \\
			takigu@kobe-u.ac.jp}
		\and
		\IEEEauthorblockN{Yasuo Ariki}
		\IEEEauthorblockA{\textit{Graduate School of System Informatics} \\
			\textit{Kobe University}\\
			Kobe, Japan \\
			ariki@kobe-u.ac.jp}
	}
	\maketitle
	
\begin{abstract}
Maintaining aging infrastructure is a challenge currently faced by local and national administrators all around the world.
An important prerequisite for efficient infrastructure maintenance is to continuously monitor  
(i.e., quantify the level of safety and reliability) the state of very large structures.
Meanwhile, computer vision has made impressive strides in recent years, mainly due to successful applications of deep learning models.
These novel progresses are allowing the automation of vision tasks, which were previously impossible to automate, 
offering promising possibilities to assist administrators in optimizing their infrastructure maintenance operations.
In this context, the IEEE 2020 global Road Damage Detection (RDD) Challenge 
is giving an opportunity for deep learning and computer vision researchers 
to get involved and help accurately track pavement damages on road networks.
This paper proposes two contributions to that topic: In a first part, we detail our solution to the RDD Challenge.
In a second part, we present our efforts in deploying our model on a local road network, 
explaining the proposed methodology and encountered challenges.		
\end{abstract}
	
\begin{IEEEkeywords}
Object Detection, Global Road Damage Detection Challenge 2020
\end{IEEEkeywords}
	
\section{Introduction}
	
In the last decade, deep learning has led to important breakthroughs in computer vision \cite{alexnet,rcnn,maskrcnn}, enabling the automation of vision tasks, which were impossible before.
The ability to automate new perception tasks is now allowing to apply quantitative approaches 
to existing problems, which were previously either economically infeasible or plainly impossible to perform quantitatively 
due to the time needed for humans to perform the visual assessment tasks.
Prominent examples of such instances in the scientific literature include the connectomics endeavor \cite{connectomics}, which requires the segmentation of neurons from petabytes of electron microscopy,
or the cataloguing of celestial objects from the visible universe \cite{celeste}. 	
Such tasks could not be approached qualitatively due to the intractable time needed by humans to identify and cross-reference objects from a massive amount of visual information.
	
Outside the scientific spheres, many industrial processes have been built around the technical constraint that perceptual tasks were not subject to automation, 
and thus limited by the time needed for humans to execute them.
One field where visual assessments at scale are of prime importance is infrastructure monitoring.
As administrations around the world face the problem of maintaining a safe network of public infrastructure,
continuously monitoring the state of this infrastructure is a global challenge of today. 
In this context, Maeda et al. \cite{rdd1} proposed a large scale dataset of Japanese road images captured from smartphone cameras, and annotated by experts with pavement damage annotations. 
In a successive project, the authors have successfully expanded the data acquisition to multiple countries, namely the  Czech Republic and India, 
to investigate the applicability and generalization of their method to different countries \cite{rdd2}.
These datasets have been made public for computer vision researchers around the world to improve the automation of pavement damage detection.

This paper consists of two distinct parts. 
In section II, we present our submission to the IEEE Big Data "Global Road Damage Detection Challenge 2020", 
providing explanations on the methods used and analyzing the failure cases of our proposed model.
In section III, we set ourselves the goal to deploy our model  in order to extract an accurate and 
exhaustive set of pavement damages on a pre-defined road network.
Our objective is to determine the feasibility and arising complications of deploying our proposed model to provide actionable insights for infrastructure managers.
We start by defining a set of goals and constraints for our deployment in section III.A, and describe the various design choices and 
challenges encountered in the rest of Section III.
Finally, in section III.G, we provide a toy problem formulation to illustrate the kind of quantitative approaches to infrastructure management enabled 
by large-scale visual intelligence extraction using computer vision deployment.

\section{Challenge}

\subsection{Dataset}

The global Road Damage Detection (RDD) Challenge dataset gathers images of road networks from Japan, India, and the Czech Republic, labeled with a bounding box and damage class annotations.
The available training set contains over $20.000$  images, of which we used $85\%$  as training data and $15\%$  as validation data.
Participants to the challenge were evaluated on two withheld test sets evaluated on the organisers' servers.
The F1 score achieved by our model on these test sets were $0.541$ and $0.543$, respectively for test set $1$ and $2$.
As the test set annotations were not made available, we report the results of our error analysis on the validation split, for which we have access to the ground truth.
For consistency, all results are presented based on this validation split.
It should be noted that our results on the validation set are higher than the results reported on the official test set.
The reasons causing this discrepancy have not been elucidated so far, but we assume that it might be due to a difference in the exact evaluation protocol.

\begin{figure}[htb]
	\centering
	\includegraphics[width=0.5\textwidth]{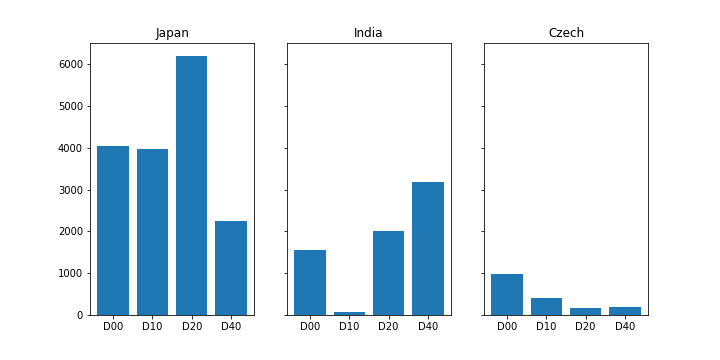}
	\caption{Distribution of the dataset damage annotations per class and countries}
	\label{dataset_dist}
\end{figure}

\subsection{Proposed method}

Our solution to the RDD challenge was based on the FasterRCNN two-stage detection architecture.
In most experiments, unless specified otherwise, we used a ResNet-50 as our baseline backbone,
the training was performed with a linear warm-up growing from 0 to $5 \times 10^{-3}$ in one epoch, followed by a cosine annealing
schedule for $20$ epochs. We used stochastic gradient descent to optimize the model with a momentum of $0.9$ and a weight decay 
parameter of  $5 \times 10^{-4}$.
The following is a list of attempts we have found to improve our results from this baseline:

\begin{itemize}
	\item Using larger backbone architectures; 
	\item Pretraining strategies; 
	\item Label smoothing (to some extent).
\end{itemize}

It is highly possible that our attempts failed due to incorrect parameterization, and that a more careful parameter search would improve our method. Nevertheless, we list our unsuccessful attempts, that brought little to no improvement to the model accuracy, below for completeness:

\begin{itemize}
	\item Learning data augmentation strategies; 
	\item Ensembling and test time augmentation; 
	\item Conditioning the network on the country information;
	\item Conditioning the bounding box regression heads on class information.	 
\end{itemize}

\begin{figure}[htbp]
\centerline{\includegraphics[width=0.9\linewidth]{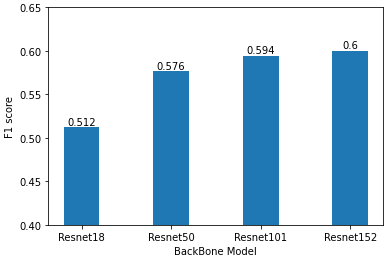}}
\caption{F1-score obtained by different backbones. Larger backbones achieve higher accuracies }
\label{backbones}
\end{figure}

\begin{figure}[htbp]
\centerline{\includegraphics[width=0.9\linewidth]{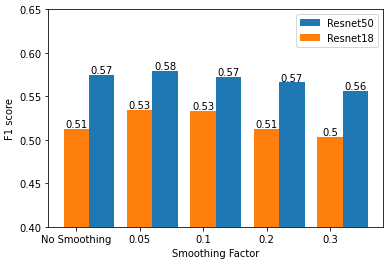}}
\caption{F1-score obtained by different parametesr of label smoothing.}
\label{labelsmooth}
\end{figure}

\begin{figure}[htbp]
\centerline{\includegraphics[width=0.9\linewidth]{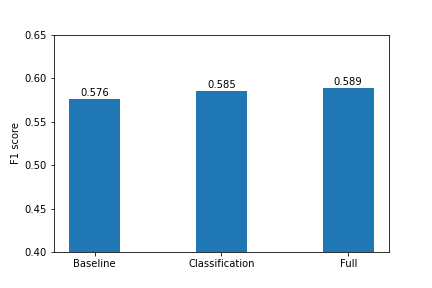}}
\caption{F1-score obtained by different pretraining strategies.}
\label{pretraining}
\end{figure}

Figure \ref{backbones} to \ref{pretraining} show the improvements obtained using each of our successful attempts.
In Figure \ref{backbones}, we show the F1 score obtained by our model using different backbones.
Increasing the bacbone model size significantly improved the results compared to the baseline resnet-50, 
although diminishing returns were observed aswe keep increasing the backbone size.

Figure \ref{labelsmooth} shows the results of our model for different label smoothing parameters.
We validate these results by running the experiment on two different backbones: Resnet-18 and Resnet-50.
A similar trend is observed with a peak F1 score obtained with a smoothing factor of 0.05. 

Finally, Figure \ref{pretraining} shows the results obtained with different backbone pre-training strategies.
The baseline pretraining strtegy corresponds to using an ImageNet pretrained backbone with randomly initialized Region Proposal Network, 
classification heads and regression heads.
The Classification pre training strategy consists in pretraining the backbone on a toy classification problem created as follows:
First we extract the inside of the damage bounding box and assign these images the label given by the dataset.
We then fine-tune an Imagenet-pretrained backbone on this artificially created dataset.
The resulting model weights were used to initialize the backbone of detection model for fine-tuning on the challenge task.
The Full strategy consisted in first training the detection model on the full available dataset, 
including all available classes (D01, D11, etc.). 
We then fine-tune the resulting model on the subset of target classes for the challenge.

\subsection{Error Analysis}

\begin{figure}[h] \centering    
	\subfigure[]{
		\label{Precision}     
		\includegraphics[width=0.4\columnwidth]{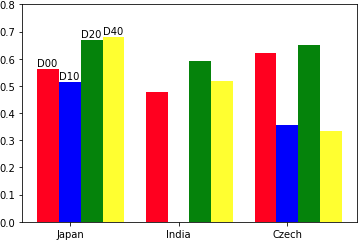}  
	}     
	\subfigure[]{ 
		\label{Recall}     
		\includegraphics[width=0.4\columnwidth]{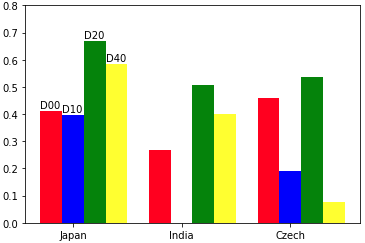}     
	}    
	\caption{Precision (a) and Recall (b) per class and countries}
	\label{prec_rec}     
\end{figure}

Figure \ref{prec_rec} shows the precision and recall per country and damage type.
The first observation to be made is that the accuracy is quite higher on images captured in Japan compared to the ones from Czech and Indian.
This difference in the results corresponds well to the imbalance of the amount of data available from the dataset, as shown in Figure \ref{dataset_dist}.

Secondly, lateral and longitudinal cracks (classes D00 and D10) tend to show lower accuracy than the other classes (D20 and D40).
One notable exception is the low level of precision and recall for the Czech dataset, which may be explained by the low amount of available data for this class (cf. Figure \ref{dataset_dist}).

Lastly, it is notable that the model could not correctly detect a single D10 instance from the Indian datset.
Similarly, this may be explained by the remarkably low amount of data available for this class.
Indeed, our validation set only contained 11 such instances.

\begin{figure}[h] \centering    
	\subfigure[]{
		\label{Tire tracks}     
		\includegraphics[width=0.45\columnwidth]{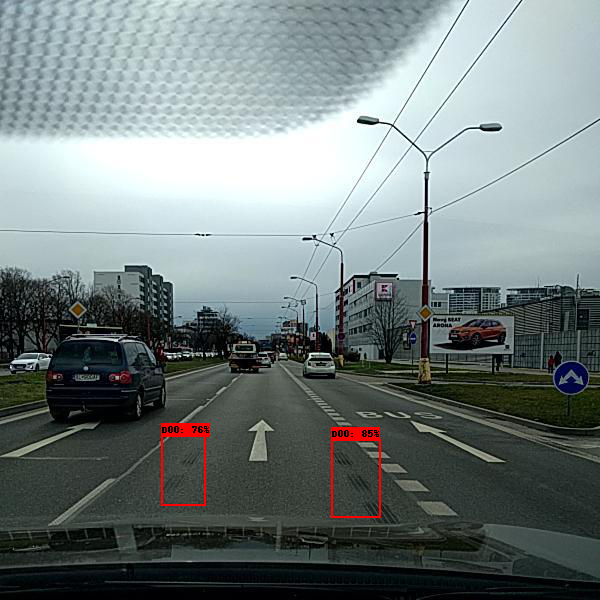}  
	}     
	\subfigure[]{ 
		\label{Shadow}     
		\includegraphics[width=0.45\columnwidth]{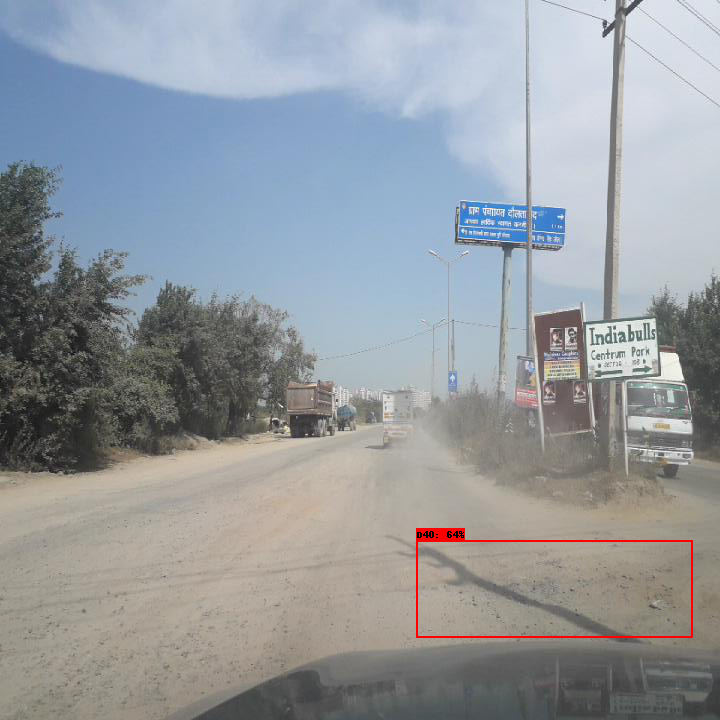}     
	}    
	\caption{ Illustrations of failure cases.  Artifacts on the roads are mistaken for damages}     
	\label{artifact}     
\end{figure}

\begin{figure}[h] \centering    
	\subfigure[]{
		\label{Clutter}     
		\includegraphics[width=0.45\columnwidth]{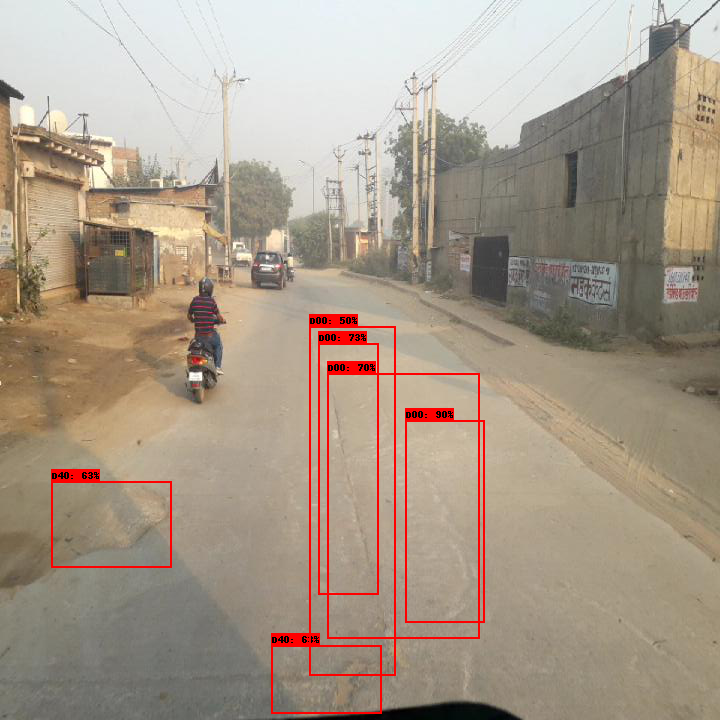}  
	}     
	\subfigure[]{ 
		\label{Full Overlap}     
		\includegraphics[width=0.45\columnwidth]{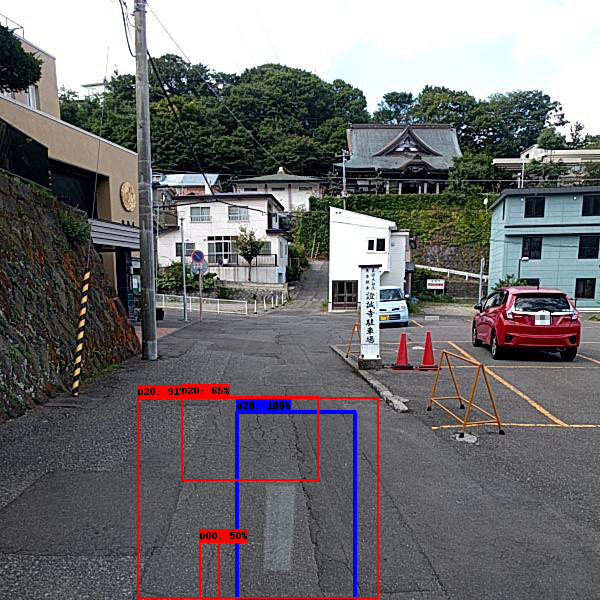}     
	}    
	\caption{ Illustrations of failure cases.  Vanilla NMS might not be best suited for crack detection. (A) cluttered detections despite NMS performed with IOU threshold of $0.5$. (B) D20 detections encompass smaller defect detections}     
	\label{NMS}     
\end{figure}

\begin{figure} \centering    
	\subfigure[]{
		\label{BD}     
		\includegraphics[width=0.45\columnwidth]{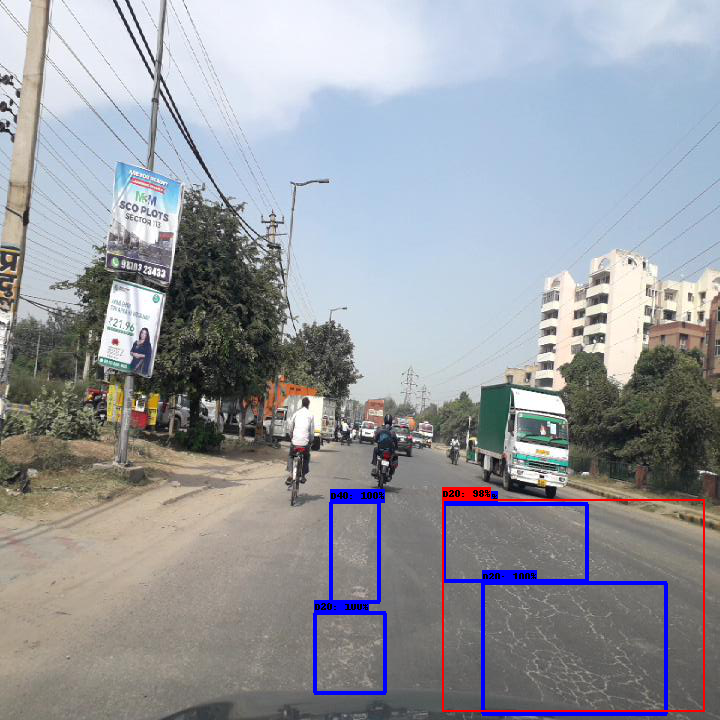}  
	}     
	\subfigure[]{ 
		\label{BGT}     
		\includegraphics[width=0.45\columnwidth]{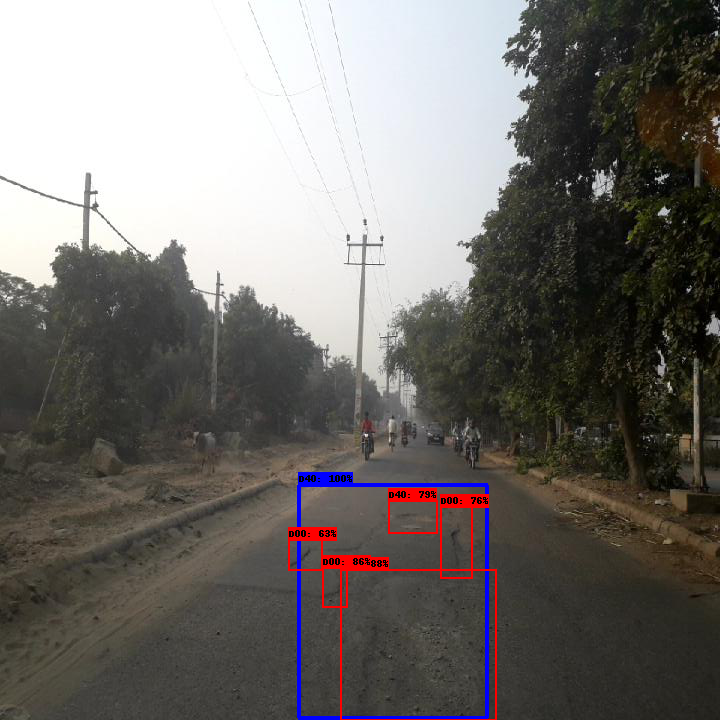}     
	}    
	\caption{ Illustrations of failure cases.  (a) Example of the model detecting of one larger area of damage and (b) the ground truth annotation defining a larger damaged area }     
	\label{exclusion}     
\end{figure}

Manual assessment of the network output revealed certain interesting phenomena that seem characteristic to pavement damage detection.
In Figure \ref{NMS} and \ref{exclusion} , we display images of the validation set annotated with ground-truth labels in blue, and model output in red.
First, a number of road artifacts, including water marks and shades, are mistaken for road damages, as illustrated in Figure \ref{artifact}     .

More interestingly, the overlap of detected damages suggest a need for a better scheme than a vanilla Non-Maximum-Suppression (NMS) scheme.
Several interesting error cases happened related to the detected damage overlap.
Figure \ref{NMS}(a) shows an example of cluttered detections. In this case, no pair of bounding box overlaps with an IOU superior to $0.5$.
Due to the very thin nature of cracks and to the coarse nature of rectangle bounding boxes, such cluttered detections may arise despite relatively low IOU thresholds.
Secondly, alligator cracks (D20) tend to cover wider areas than the other classes of damages.
Within these wider ranges, the model tends to assign additional detections to subsets of these spaces.
These kinds of errors might be dealt with by hand-crafted rules excluding detections contained within larger findings.

Finally, closely related to the phenomenon discussed above, a frequent pattern of errors consisted in assigning either too few or too many distinct detections to areas of neighboring damages.
Figure \ref{exclusion}  illustrates such error cases. In Figure \ref{exclusion} (b), the ground-truth data assigns a single damage bounding box to the area, 
whereas the model distinguishes between different cracks of this area as different damages.
In Figure \ref{exclusion} (a), the inverse situation happens in which our model assigns a single damage to the wider damaged area, 
whereas ground-truth annotations assign finer-grained detections.
To our novice eyes, and without knowledge of rules clearly defining the separation between neighboring damages, 
both outputs seem acceptable and the quantitative accuracy seems to be left to the annotator's appreciation.
	
\section{Deployment}
	
\subsection{Motivation and goals}

In this section, we investigate the feasibility, and estimate the difficulties of, extracting 
an accurate and exhaustive set of geo-localized pavement defects of a target road network.
We ask ourselves the following questions: what engineering challenges lie in-between running our model inference
and providing actionable insights to infrastructure managers?
To uncover these challenges, we set ourselves the following goals:

\begin{itemize}
	\item To extract an exhaustive dataset of geo-localized pavement damages on a given target road network; 
	\item At minimum cost; 
	\item To illustrate the usability of this dataset in developing quantitative approaches to planning maintenance operations.
\end{itemize}

Realizing these goals requires answering the following questions:

\begin{itemize}
	\item How can one efficiently image a target road network in its entirety? 
	\item How can one deal with errors and uncertainty in the model output? 
	\item How can one minimize the cost of such endeavor?
\end{itemize}

We have made the following design decisions to address the above questions:

\begin{itemize}
	\item Centralize and optimize the image acquisition process.
	\item Provide a geo-localized interface to visualize the extracted damages, allowing users to manually assess and correct the model errors.
	\item Rely exclusively on low-cost hardware, open data and software.  
\end{itemize}
	
\subsection{Deployment process}
	
\begin{figure}[htbp]
\centerline{\includegraphics[width=0.9\linewidth]{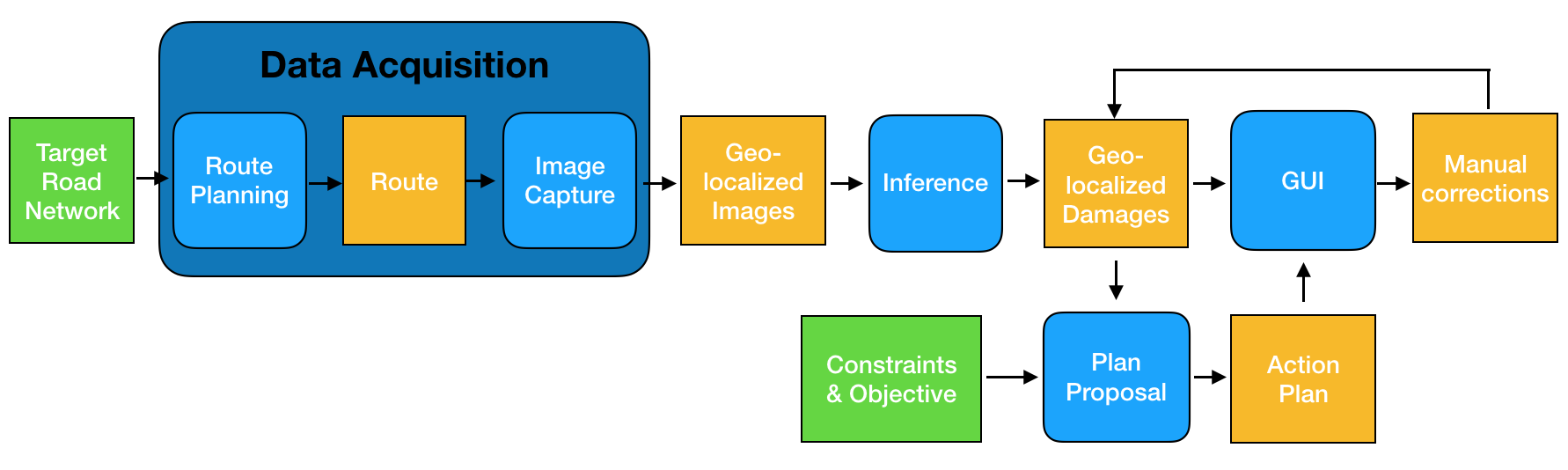}}
\caption{Illustration of the proposed methodology}
\label{service}
\end{figure}

Figure \ref{service} illustrates the architecture of the proposed method.
First, the data acquisition phase generates a set of geo-localized images of the given road network.
We further subdivide the task of data acquisition into the two sub-tasks of route planning and image capture.
Second, images are processed by our model to extract geo-localized pavement damages.
These defects are then integrated to a GUI interface allowing the user to browse his or her section of interest,
assess the validity of extracted damages and fix the potential model errors.

Finally, we propose a toy problem formulation expressing the challenge of finding an optimal maintenance operation as a constrained optimization problem over the extracted dataset.
This proposition is meant as an illustration of the idea presented in the introduction:
visual intelligence gathered at scale using computer vision opens the way to new quantitative approaches to solve existing problems. 

The following sections describe the progresses made and challenges encountered within each of these steps.
		
\subsection{Route planning}

\begin{figure}[htbp]
\centerline{\includegraphics[width=0.9\linewidth]{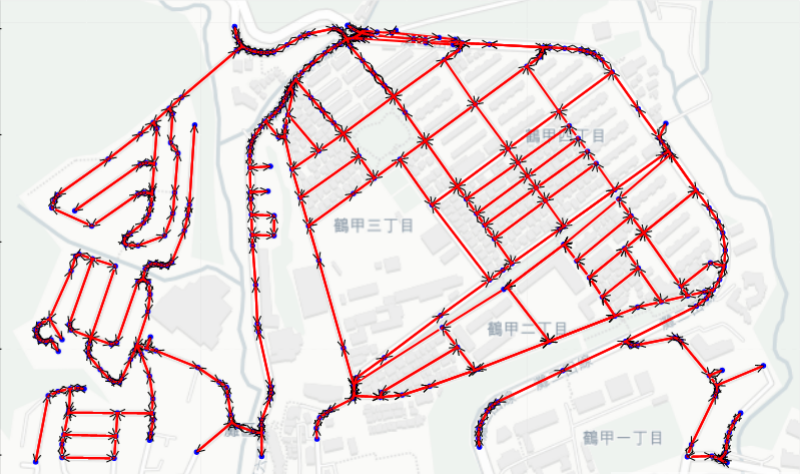}}
\caption{Directed graph extracted from OSM data representing our target road network }
\label{road_net}
\end{figure}

The first step of our method consists in defining a route for the vehicle to follow, spanning the entirety of the target road network. 
To compute this route, we relied on Open Street Map data \cite{osm}, which we downloaded and preprocessed using the OSMNX library \cite{osmnx}.
The result of this preprocessing is a directed graph $\mathcal{G} = \big\{E,V\big\}$ in which vertices $v_i \in V$ represent road intersections, and edges $e_i \in E$ serve as roads between these intersections.
Each edge is annotated with a distance in meters $d(e_i),  \forall e_i \in E$.
The directed graph extracted for our target road network is illustrated in Figure \ref{road_net}.

Given the extracted graph, our goal of finding a route to efficiently image the road network in its entirety can be formulated as follows:
Find a path $P= \big( e_{0} \rightarrow  e_{1} \rightarrow ... \rightarrow e_{N} \big) $ of minimal distance $d(P)= \sum_{e_j \in P} d(e_j)$ traversing the whole set of edges so that $E \subset P$ at least once.
This problem definition corresponds to the well-known Chinese Postman Problem \cite{cpp} which admits a fairly simple solution in two steps:
graph eulerization and Euler circuit generation.

For the eulerization step, we implemented the scheme presented in \cite{cppsolpaper}, using the CVXPY library \cite{cvxpy} to offload calls to the ECOS \cite{solver} integer program solver.
The resulting eulerized graph has a total length of 28.0km, which represents an addition of 6.9\% or road navigation compared to the original road network of 26.2 km.

For the Euler circuit generation, we implemented a variant of the Fleury algorithm, augmented with a priority queue to minimize the amount of U-turns and left turns in the generated circuit.

We then converted the resulting circuit to a GPX navigation file and uploaded it to a smartphone.
We used the OsmAnd open-source application to get real-time navigation inctrustions.

Although this approach allowed for an effective road network navigation, several concerns remain:
One issue, for example, was the imaging of multi-lane streets: our setup currently only allows the imaging of a single lane. 
Integrating lane information to the real-time navigation system would be quite challenging.
Another limitation of our approach is that it assumes a constant driving speed on all roads and intersections, and ignores traffic conditions and traffic signals completely (traffic lights, stop signs, pedestrians, etc.).
Relying on specialized navigation software would probably improve the navigation efficiency, but no open solution fitting our needs was found.

\subsection{Image Capture}

Bearing our goal of a modest acquisition cost in mind, we used a RaspBerry Pi 4 as a computing platform, controlling a PiCamera HQ camera for image capture. 
We connected a 500GB external SSD storage to store the captured images and a GPS receiver for geo-localization.
For reasons of simplicity and practicability, a personal scooter was employed as the driving vehicle whereby the camera was fixed to the front end through an articulated arm. 
The entire setup was created under \$200.

Each picture was taken with a resolution of 2032 by 1520 pixels at a recording frame rate of 10 images per second.
Given the maximum speed limit of 40 km/h on the target road network, at least one photo was taken every 1.5 meters.
With these specifications, our setup would allow for a continuous recording around 385.000 images during a time window 
of more than 10 hours before exceeding the capacity of the hard drive. 

Each recorded image was indexed by the time stamp of its capture.
The GPS receiver generates a set of coordinates similarly indexed by the time stamp of their reception. 
Images were then mapped to a unique coordinate by linearly interpolating their time stamp against the GPS time stamp series.

Finally, we associate each image to an edge in the road network graph.
We do so by computing the distance between the image coordinates to each of the network edges.
For simplicity, and given the short distances considered, the point to segment distances was computed under a planar assumption.

\begin{figure}[htbp]
\centerline{\includegraphics[width=1.1\linewidth]{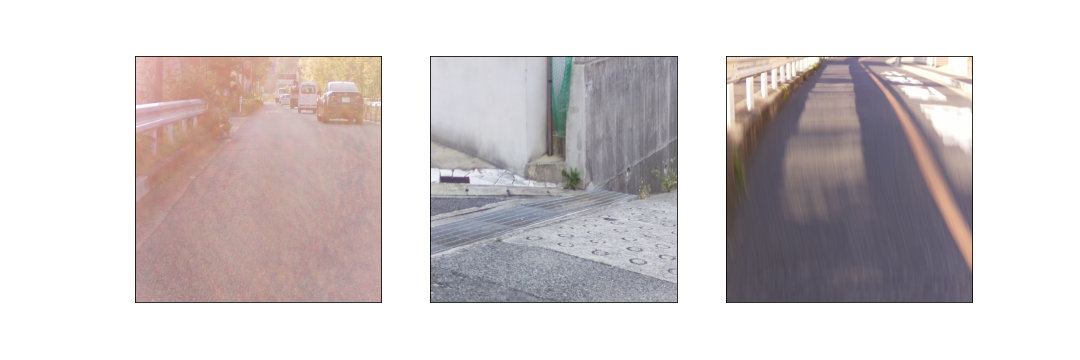}}
\caption{Illustration of the noise in image captures. (left) Illumination noise. (center) Sharp corners yield walls and sidewalks captures. (right) Motion blur.}
\label{imnoise}
\end{figure}

A few hurdles we encountered while capturing the images are worth mentioning.
The most problematic one was the vibrations transmitted from the vehicle to the camera.
As a result, several pictures suffer a significant motion blur, as illustrated in Figure \ref{imnoise}.
Second, illumination variations also affected the quality of the captured images.
In particular, moving from shaded areas to high illumination areas, the camera processing pipeline 
had trouble adjusting to the new illumination of the environment. 
These problems deserve attention and might be fixed by a better calibration of the camera.
Finally, as the target road network consists mostly of small roads, several sharp corner turns were made.
In these cases, the camera setup ended up capturing mostly walls and sidewalks, as illustrated in Figure \ref{imnoise}.
Such captures should be filtered to avoid detecting wall cracks or other undesired artifacts.

\subsection{Inference}

Once the geo-localized images acquired, we fed them to the model described in Section 2 to extract a (possibly empty) set of damages identified by the output label of the damage $l$, 
its location within the image as a bounding box $b$, and a confidence score $s$ ranging from $0$ to $1$.
We then associate to each extracted defects the geo-localization of their origin image.

This approach raises two (currently unsolved) problems:
First, the geo-localization of the damages does not correspond to their actual coordinates, 
but to the coordinates of a point of view from which they are visible.
This may result in detected damages to be assigned to the wrong edge of the network.
Several solutions to this problem may be considered: 
One might either refine the geo-localization by combining the GPS information with the content of the image.
However, this would be non-trivial to realize.
Alternatively, one might design a capture setting with a narrower field of view (e.g., a top down point of view) and higher frame rates. 
This would probably necessitate a new dataset adapted to this point of view.

Second, many damages are visible from different images, leading to duplicate detections of the same damage.
We distinguish two kinds of duplicates:

Some duplicates are due to the immobility of the data capture at a given time (due to traffic lights, stops, or turns).
To deal with these duplicates, we sub-sampled the set of target images using their coordinates, selecting only one image every ten meters.
A Gaussian smoothing of the GPS time series was performed before subsampling in order to smooth the GPS noise.

\begin{figure} \centering    
	\subfigure[]{
		\label{BD}     
		\includegraphics[width=0.45\columnwidth]{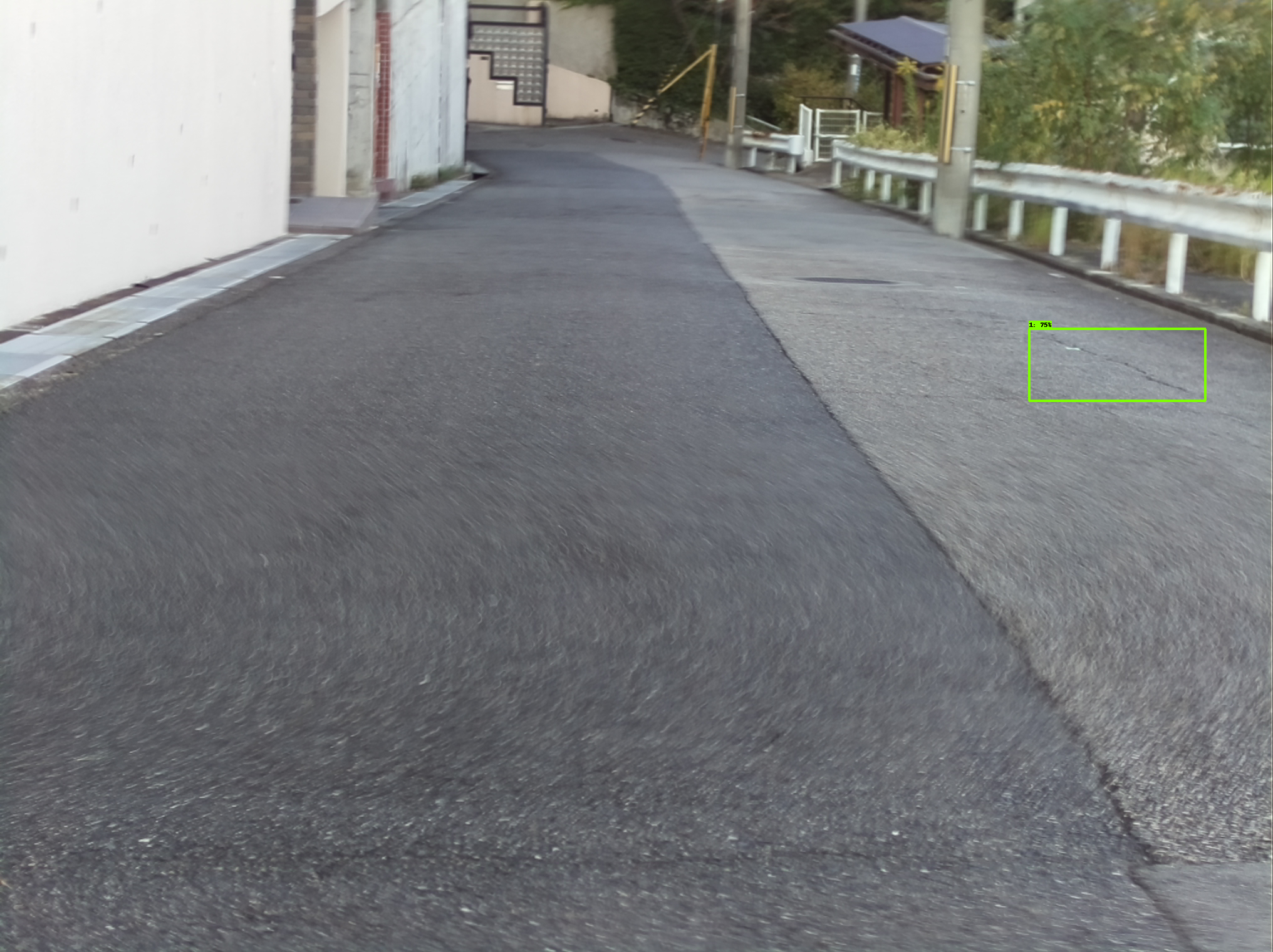}  
	}     
	\subfigure[]{ 
		\label{BGT}     
		\includegraphics[width=0.45\columnwidth]{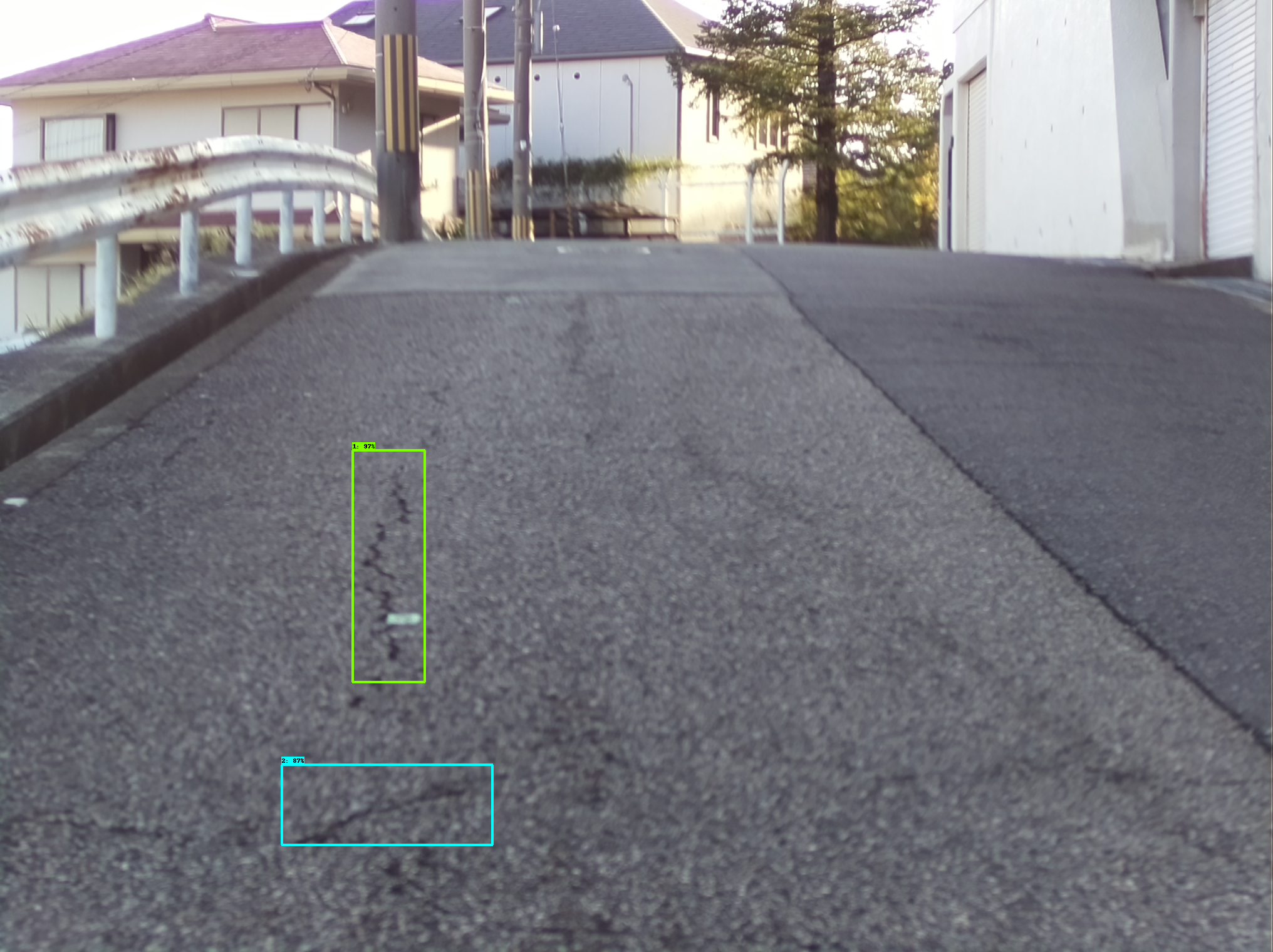}     
	}    
	\caption{ Illustration of duplicate detections }     
	\label{dup}     
\end{figure}

Other duplicates correspond to the same location being imaged from different directions or intersection ways.
These instances are harder to deal with and have not yet been dealt with at the time of this writing.
Figure \ref{dup} illustrates one instance of such duplicate detection. 

\begin{figure} \centering    
	\subfigure[]{
		\label{BD}     
		\includegraphics[width=0.45\columnwidth]{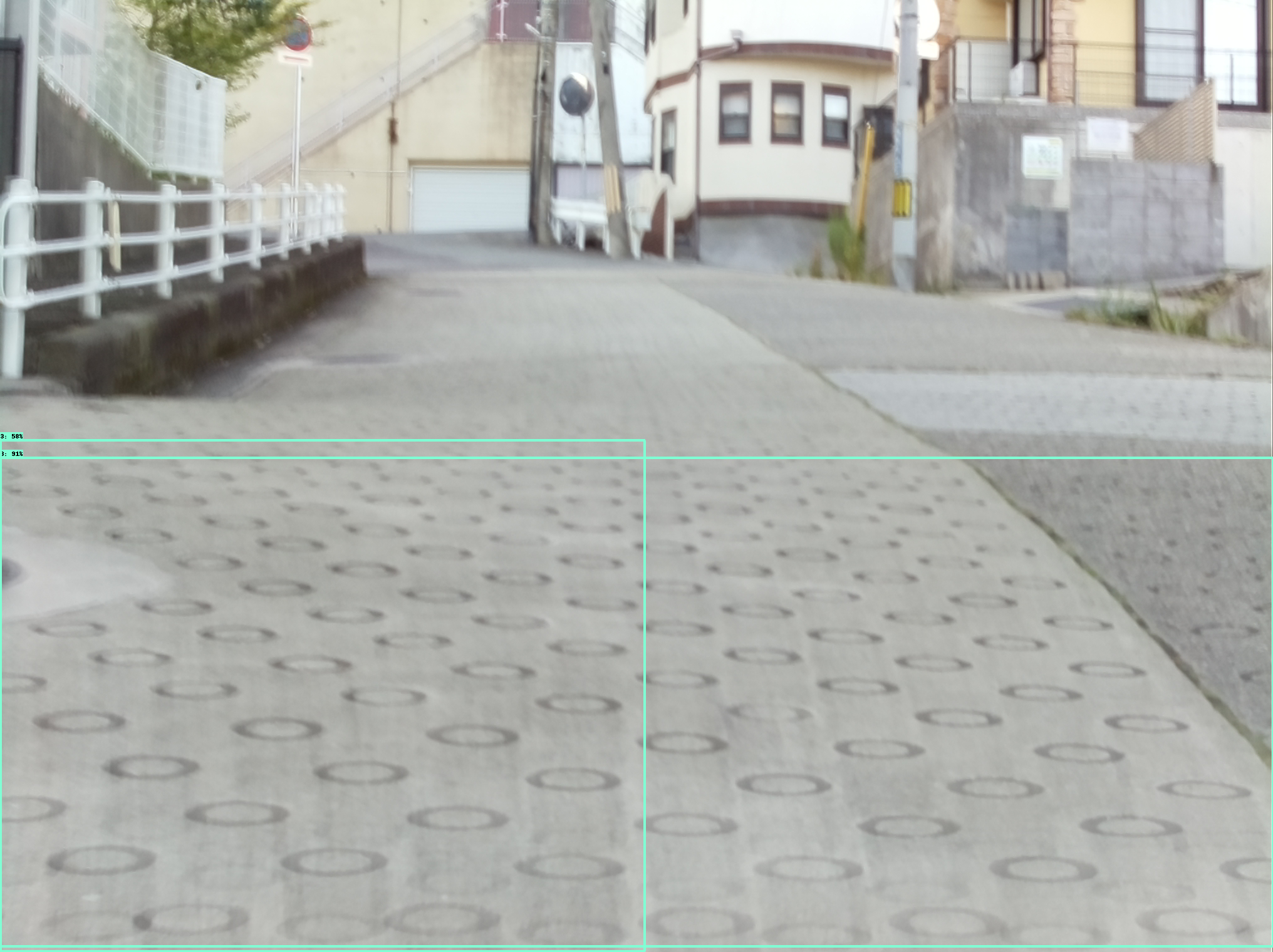}  
	}     
	\subfigure[]{ 
		\label{BGT}     
		\includegraphics[width=0.45\columnwidth]{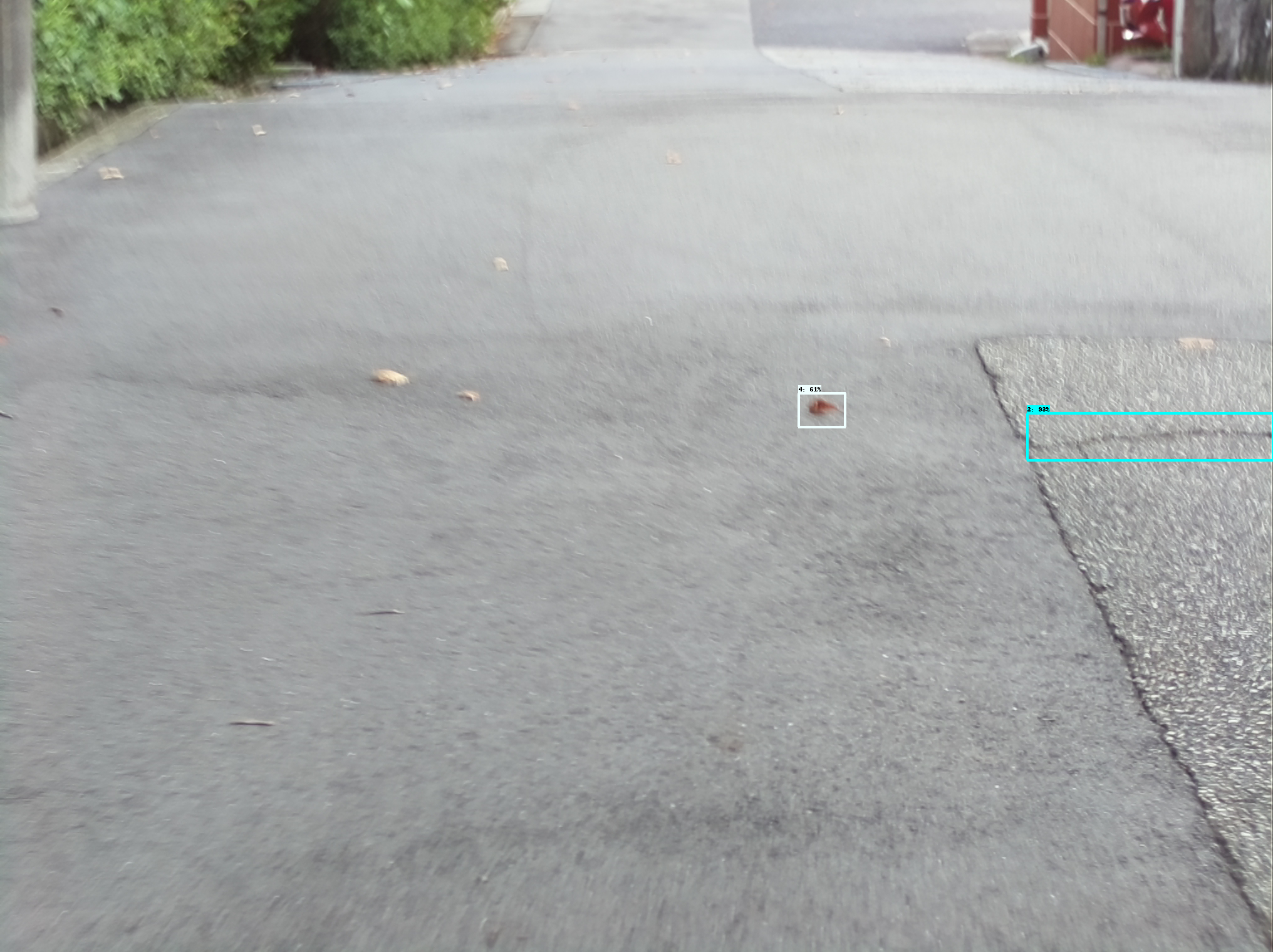}     
	}    
	\caption{ Illustration of domain shift complications (a) poor generalization to new pavement types (b) dead leaves are detected as damages, showing that different seasonal features may impact generalization.}     
	\label{miss}     
\end{figure}

Other failure cases that may be attributed to domain shift are illustrated in Figure \ref{miss}. 
Notably, we notice a kind of pavement that triggers the model to systematically detect non-existing alligator cracks (Figure \ref{miss} (a)),
and several dead leaves being mistaken by our model for pothole-kind of damages (Figure \ref{miss} (b)).

\subsection{Graphical User Interface}

\begin{figure}[htbp]
\centerline{\includegraphics[width=0.9\linewidth]{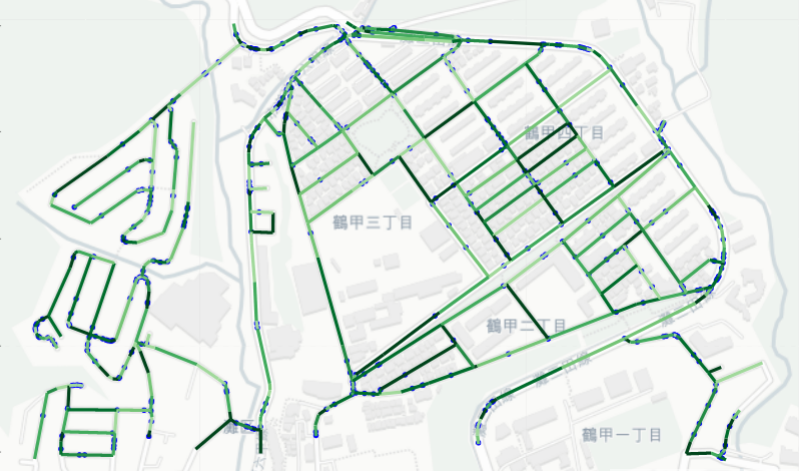}}
\caption{GUI Illustration: road segment visualization}
\label{gui1}
\end{figure}

\begin{figure}[htbp]
\centerline{\includegraphics[width=0.9\linewidth]{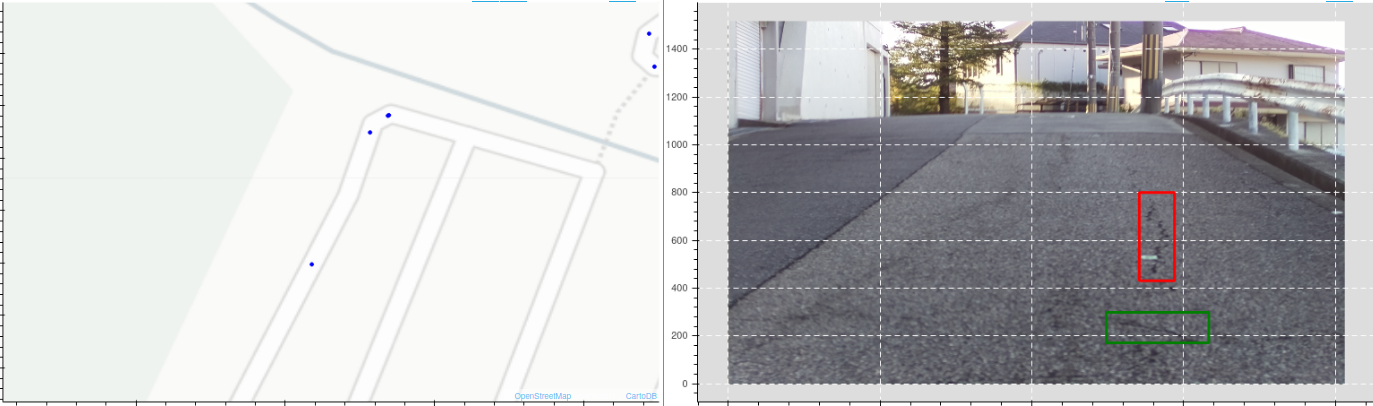}}
\caption{GUI Illustration: Manual inspection of extracted damages}
\label{gui2}
\end{figure}

Figure \ref{gui1} and \ref{gui2} show snapshots of our proposed GUI.
The first interface, represented in Figure \ref{gui1}, lets users interactively browse a map of the network, 
displaying road edges in a color scale representing the extracted damage severities.
In its current state, the damage severity scale represents the sum of detected damage probabilities per unit of distance on the edge.
However, more pertinent measures of road conditions should be considered for practical applications.

Figure \ref{gui2} shows the second interface, displaying extracted damages as geo-localized points on the map.
Selecting a point on the map displays the associated image.
This interface is intended to let users manually verify and correct the extracted damages.  

\subsection{Action plan optimization}

In this section, we propose a toy optimization problem simulating the task of planning for optimal maintenance operations.
This toy problem is meant as an example of quantitative approaches 
enabled by the large-scale extraction of visual information to illustrate the idea presented in the introduction, 
rather than an attempt to tackle an actually practical use-case.  
Practical use-cases would require much more complex constraint and objective definitions.

Let us define a cost function $c(e)$ over a single edge $e \in E$ of the road network as the sum of the damage probabilites extracted along this edge. 
In addition, each edge $e$ is associated a time $t(e)$, modeling the time required by a maintenance operation on this edge.
We augment the graph by inserting duplicate edges $e'$ for all edges $e \in E$ of the graph.
Duplicate edges are annotated with a time $t(e')=s \times d(e)$ and cost $c(e')=0$ to represent the action of traversing the edge without maintenance. 
Here $s$ represent a navigation speed constant converting distances to times.

Finally, we define a maintenance plan as a path $P \in E_{aug}$ starting from a root node $u_0 \in V$ in this augmenting graph.
This path is intended to represent a daily maintenance deployment from an agent based at root node $u_0$,
traversing the road network graph, selectively maintaining a subset of the edges traversed.
We define a path cost $c(P)$ and time $t(P)$ as the sum of the respective values of its edges:

\begin{equation}
\begin{aligned}
& c(P)=\sum_{e \in P} c(e) \\
& t(P)=\sum_{e \in P} t(e)
\end{aligned}
\end{equation}

Given the above definitions, the problem of finding an optimal maintenance operation under a given time budget $T$ can be formulated as follows:
Find a path $P \in E_{aug}$ with maximal cost $c(P)$, 
under the constraint that $t(P) < T$, 
and that none of the original network edges from $E$ can be crossed more than once (i.e., the agent can only fix a road segment once).

For small enough networks, simple procedural solutions can be implemented.
To deal with larger networks, or integrate more realistic constraints, more advanced solutions would need to be considered.



\subsection{Limitation and Future Work}

Much work remains to be done in order to propose a practically useful system. 
In this subsection, we detail the most important limitations of the presented work.

\subsubsection{Current implementation limitations} 
We see two major shortcomings of the proposed approach in its current form. 
First, the captured images contain a lot of noise, due to motion blur and illuminations.
Although we have not quantified drops in accuracy due to this noise, we suspect this has an important impact on the completeness of the extracted road damages.
This also negatively affects the GUI experience.
Better imaging could be achieved with better handling of the vehicle vibrations and better calibration of the camera.  
Second, we have not yet dealt with the problem of duplicate detections of damages. Future work should focus on this important problem. 

\subsubsection{Current design limitations} 
The current design of the proposed approach has several important flaws.
First, the severity of the extracted damages is not assessed. 
We have relied on the detection score of the detection outputs to assess the importance of the detected damages. 
We believe this to be an important shortcoming of the current system as we expect damage severity to be one of the most important insight 
to drive subsequent maintenance operation decisions.
Assessing damage severity might be done by integration of such techniques as crack segmentation \cite{crack_seg} or pothole classification \cite{pothole}. 
Finally, our current approach does not quantify, nor deal with the uncertainty in the geo-localization of the damages. 
Most notably, the current geo-localization of the extracted damages does not reflect their true position, 
but rather the position of a point of view from which the damages are visible.
This leads, in some cases, to damages being associated to the wrong segment of the road network.
At the time of this writing, no viable solution to this problem has been considered.

\section{Conclusion}

The recent success of computer vision is enabling the automation of previously impossible visual tasks.
One sector of activity where this automation is set to have a profound impact is infrastructure management where continuous monitoring of
large-scale infrastructures is required.
Building on excellent prior works \cite{rdd1,rdd2} in this direction, this paper has detailed our solution to the Global Road Damage Detection Challenge 2020, 
as well as our efforts to deploy the resulting model into a functional service aiming to assist infrastructure managers.
In our efforts, we point out several challenges related to deployment.

\section*{Acknowledgment}

This work was supported by JSPS KAKENHI Grant Numbers JP20K19823, JP19K24344

\vspace{12pt}
\end{document}